%% file: arxiv_main.tex
\documentclass{article}

\usepackage{arxiv}
\usepackage{amsmath}
\usepackage[utf8]{inputenc} 
\usepackage[T1]{fontenc}    
\usepackage{hyperref}       
\usepackage{url}            
\usepackage{booktabs}       
\usepackage{amsfonts}       
\usepackage{nicefrac}       
\usepackage{microtype}      
\usepackage{cleveref}       
\usepackage{graphicx}
\usepackage{doi}
\usepackage{algorithm}
\usepackage{algorithmic}

\usepackage{subfig}
\usepackage{siunitx}
\usepackage{amssymb}
\usepackage{bbding}
\usepackage{ragged2e} 
\usepackage{makecell, multirow, tabularx}
\usepackage{wrapfig}

\title{SocialCVAE: Predicting Pedestrian Trajectory via Interaction Conditioned Latents}

\author{
  \textbf{Wei Xiang}$^{1}$, \textbf{Haoteng Yin}$^{2}$, \textbf{He Wang}$^{3}$, \textbf{Xiaogang Jin}$^{1}$
  \\
  $^1$ State Key Lab of CAD\&CG, Zhejiang University \quad $^2$ Department of Computer Science, Purdue University \\
  $^3$ Department of Computer Science, University College London \\
  \texttt{xiangwvivi@gmail.com, yinht@purdue.edu, he\_wang@ucl.ac.uk, jin@cad.zju.edu.cn}
}

\usepackage{xspace}
\newcommand{\proj}{SocialCVAE\xspace}
\hypersetup{
pdftitle={A template for the arxiv style},
pdfsubject={q-bio.NC, q-bio.QM},
pdfauthor={David S.~Hippocampus, Elias D.~Striatum},
pdfkeywords={First keyword, Second keyword, More},
}

\input{math_commends}

\begin{document}
\maketitle

\begin{abstract}
Pedestrian trajectory prediction is the key technology in many applications for providing insights into human behavior and anticipating human future motions. Most existing empirical models are explicitly formulated by observed human behaviors using explicable mathematical terms with a deterministic nature, while recent work has focused on developing hybrid models combined with learning-based techniques for powerful expressiveness while maintaining explainability. However, the deterministic nature of the learned steering behaviors from the empirical models limits the models' practical performance. To address this issue, this work proposes the social conditional variational autoencoder (\proj) for predicting pedestrian trajectories, which employs a CVAE to explore behavioral uncertainty in human motion decisions. SocialCVAE learns socially reasonable motion randomness by utilizing a socially explainable interaction energy map as the CVAE's condition, which illustrates the future occupancy of each pedestrian's local neighborhood area. The energy map is generated using an energy-based interaction model, which anticipates the energy cost (i.e., repulsion intensity) of pedestrians' interactions with neighbors. Experimental results on two public benchmarks including 25 scenes demonstrate that \proj significantly improves prediction accuracy compared with the state-of-the-art methods, with up to 16.85\% improvement in Average Displacement Error (ADE) and 69.18\% improvement in Final Displacement Error (FDE).
Code is available at: \url{https://github.com/ViviXiang/SocialCVAE}.
\end{abstract}

\input{0-Introduction}
\input{1-RelatedWork}
\input{2-Methodology}
\input{3-Evaluation}

\input{4-Conclusion}
\input{6-acknowledgement}
\bibliographystyle{ACM}
\bibliography{reference}

\newpage
\appendix
\input{5-Appendix}

\end{document}

%% file: math_commends.tex
\usepackage{amsmath,amsfonts,bm}

\def\eqref#1{equation~\ref{#1}}

\def\1{\bm{1}}

\def\rh{{\textnormal{h}}}

\DeclareMathAlphabet{\mathsfit}{\encodingdefault}{\sfdefault}{m}{sl}
\SetMathAlphabet{\mathsfit}{bold}{\encodingdefault}{\sfdefault}{bx}{n}
\newcommand{\tens}[1]{\bm{\mathsfit{#1}}}

\newcommand{\etens}[1]{\mathsfit{#1}}



%% file: 0-Introduction.tex
\section{Introduction}
Pedestrian trajectory prediction is a vital task in intelligent systems for understanding human behavior and anticipating future motions. Predicting the future movements of pedestrians in complex environments is challenging due to the highly dynamic and subtle nature of human interactions.

Empirical methods explicitly model interactions for crowd motion prediction, e.g., rule-based model \cite{reynolds1987boids,reynolds1999steering}, force-based model \cite{helbing1995socialforce, karamouzas2014powerlaw} and energy-based model \cite{guy2010pledestrians,karamouzas2017implicit}. These models are explainable but with lower predictive accuracy, as they cannot fit observed data precisely.
 
In contrast, various methods based on deep neural nets have been proposed with social interaction modeling by employing social pooling mechanism \cite{alahi2016sociallstm, gupta2018sgan}, graph-based modeling \cite{mohamed2020socialstgcnn,bae2021disentangled}, and attention mechanism \cite{mangalam2020pecnet, shi2021sgcn, tsao2022socialssl}. While they achieve expressive power and generalization ability, their black-box nature makes the learned model less interpretable to human understanding. It remains a challenge to explore the trade-off between model explainability and prediction capability. Recent research effort has been focused on exploring the aforementioned trade-off by designing hybrid models that combine deep neural nets with explainable interaction \cite{kothari2021socialanchor, yue2022nsp}. However, their prediction accuracy suffers from the deterministic nature of the physics-driven behaviors~\cite{yue2023human}.

\begin{figure*}[t]
    \centering
    \includegraphics[width=\linewidth]{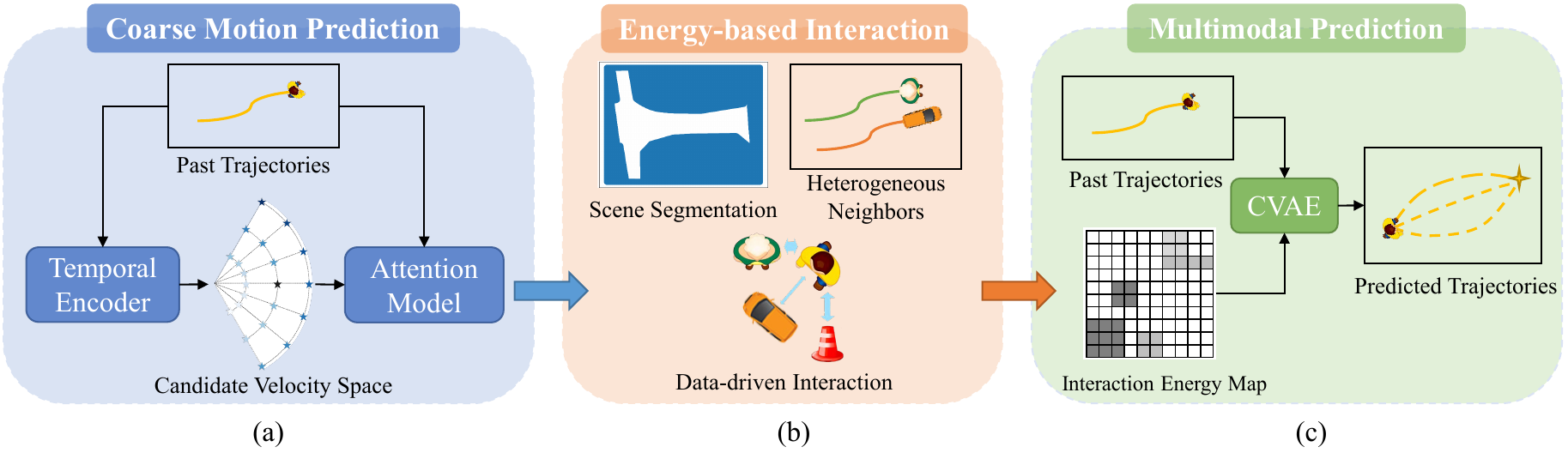}
    \caption{The framework of \proj. (a) The coarse motion prediction model learns the temporal motion tendencies and predicts a preferred new velocity for each pedestrian. (b) The energy-based interaction model constructs a local interaction energy map to anticipate the cost of pedestrian interactions with heterogeneous neighbors, including pedestrians, static environmental obstacles found in the scene segmentation (e.g., buildings), and dynamic environmental obstacles (e.g., vehicles). (c) The multimodal prediction model predicts future trajectories using a CVAE model conditioning on the past trajectories and the interaction energy map.
    }
    \label{fig:framework}
\end{figure*}

To overcome the challenges while retaining the advantages of hybrid methods, we propose \proj, a new hybrid model for pedestrian trajectory prediction that combines an energy-based interaction model for socially explainable interaction anticipations with an interaction-conditioned CVAE for multimodal prediction.
Fig. \ref{fig:framework} illustrates the framework of our method.
\proj takes advantage of the data-driven optimization model \cite{xiang2023model} to quantify the interaction energy cost (i.e., repulsion intensity) of the temporal coarse predictions and explicitly represent the interaction energies into the local energy map. Using the CVAE model conditioned on the interaction energy map, \proj learns socially reasonable residuals for the temporal motion decisions. Similar to the previous methods \cite{zhou2021scsr, yue2022nsp} that achieve state-of-the-art (SOTA) performance, we employ the recursive prediction scheme to update future trajectories step by step with the input trajectories at each step including the updated trajectories. 

We conduct extensive experiments on two popular benchmark datasets (ETH-UCY \cite{pellegrini2009you, lerner2007crowds} and SDD \cite{robicquet2016learning}), and demonstrate \proj's superiority over existing state-of-the-art methods in terms of prediction accuracy. Furthermore, our results highlight the effectiveness of using an energy-based interaction model for pedestrian trajectory prediction and provide insights into how to better model pedestrian behavior in complex environments.
The main contributions are concluded as follows:
\begin{itemize}
    \item We propose a novel multimodal pedestrian trajectory prediction model (\proj) that leverages the advantages of both empirical and learning-based approaches for better prediction performance and interpretability of motion decisions.
    \item \proj explores the behavioral uncertainty of human motion by introducing socially explainable interaction energy maps generated from an energy-based interaction model. Both the quantitative and quality results of \proj demonstrate that the energy-based interaction helps the model better understand the social relationships between pedestrians, leading to improved prediction performance.
\end{itemize}

%% file: 1-RelatedWork.tex
\section{Related Works}
\subsection{Energy-Based Interaction Methods}
Considering the nonlinear nature of pedestrian motion dynamics that pedestrians try to anticipate and react to the future trajectories of their neighbors for collision avoidance \cite{karamouzas2014powerlaw}, 
energy-based methods \cite{karamouzas2017implicit, ren2019heter, xiang2023model} predicts pedestrians' future trajectories by minimizing the anticipated social interaction cost calculated by energy functions, i.e., the anticipated repulsion intensity from neighbors. These models explicitly predict pedestrians' future motion by consuming minimum interaction cost, but provide less prediction accuracy due to relying solely on explicit motion features such as velocity. Comparatively, our method is a hybrid model that leverages the social explainability of energy-based interaction models with the prediction capability of deep-learning models, resulting in better prediction performance.

\subsection{Data-Driven Methods}
With advances in data acquisition techniques, deep learning methods have been proposed and have achieved impressive results in predicting pedestrian trajectories.
RNN structure has widely been used to capture temporal dependencies while considering social interactions using pooling mechanism \cite{alahi2016sociallstm, bisagno2018grouplstm, gupta2018sgan} or attention mechanism \cite{vemula2018socialattention, sadeghian2019sophie,salzmann2020trajectron++,xu2022socialvae}.
Graph-based models that utilize distance-based physical adjacency matrices \cite{mohamed2020socialstgcnn,bae2021disentangled,xu2022groupnet} or attention-based learnable adjacency matrices \cite{huang2019stgat,shi2021sgcn,duan2022cagn,wu2023msrl} to learn pedestrian social interactions have also been developed. Besides, transformer-based models incorporate attention mechanisms \cite{yu2020star,yuan2021agentformer,tsao2022socialssl} to model social interaction for better performance in pedestrian trajectory prediction tasks.

Recently, prediction accuracy improvement has been made by NSP-SFM \cite{yue2022nsp}, a multimodal prediction model which is a hybrid of steering behavior learning based on conservative position-dependent forces with unexplainable randomness learning.
However, the deterministic force-driven behavior of NSP may result in performance degradation~\cite{yue2023human}.
Different from NSP-SFM, our method combines the energy-based interaction model for explicit interaction cost anticipation with interaction-conditioned human motion uncertainty learning, resulting in providing socially reasonable randomness of future motion and yielding superior prediction performance.

%% file: 2-Methodology.tex
\section{Methodology}
\subsection{Problem Formulation}
Pedestrian trajectory prediction aims to predict the positions of pedestrians' trajectories in a traffic scenario. Given the observed $T$-time step pedestrian trajectories $\mathcal{X}_{1:T} = \{ X_1, X_2, \ldots, X_{T} \}$, the task is to predict the pedestrians’ future trajectories $\hat{\mathcal{X}}_{T+1:T+M}=\{\hat X_{T+1}, \hat X_{T+2}, \ldots, \hat X_{T+M}\}$ over the next $M$ time steps, where $X_i=\{\mathbf{x}_1^t, \dots, \mathbf{x}_n^{t} \} \in \mathbb{R}^{n \times 2}$ denotes the spatial (2D-Cartesian) coordinates of $n$ pedestrian at time step $t$. Formally, taking the past $T$-time step trajectories of pedestrians $\mathcal{X}_{t-T+1:t}$, the trajectories of other dynamically moving obstacles (e.g., vehicles) $\mathcal X_{d}$, and the scene segmentation $S$ of the surrounding environment (see Fig. \ref{fig:framework}b, the colored area is not feasible for human motions) as input, the prediction task is formulated as:
\begin{equation}
    \mathcal{\hat{X}}_{t+1} = f(\mathcal{X}_{t-T+1:t}, \mathcal{X}_d, S),
\end{equation}
where $f$ is any prediction model.

\subsection{Framework}
The framework of our method is illustrated in Fig. \ref{fig:framework}. Overall, \proj learns the uncertainty of human motion and implements it as predicting residuals of a coarse prediction.
Specifically, a coarse prediction model (Fig. \ref{fig:framework}a) predicts a temporally reasonable preferred velocity and a new position for each pedestrian by aggregating the information from a discrete candidate velocity space, which is built based on the learned temporal tendency from an RNN-structured temporal encoder and includes possible temporal motion decisions (velocities). 
Then, an energy-based interaction model (Fig. \ref{fig:framework}b) anticipates the social interaction cost of the preferred velocity with heterogeneous neighbors, including interactions with pedestrians, static obstacles (e.g., buildings) obtained from the scene segmentation, and dynamic obstacles (e.g., vehicles). We map the interaction energy with the neighbors onto a local energy map to represent the future occupancy of the local neighborhood area.
Finally, a CVAE model (Fig. \ref{fig:framework}c), which is conditioned on both past trajectories and the interaction energy map, predicts the socially reasonable residuals of the preferred new position and generates multimodal future trajectories.

\subsection{Coarse Motion Prediction}
\label{subsec:coarseprediction}
The coarse motion prediction model predicts a temporally reasonable future motion (velocity $\mathbf{v}$, position $\mathbf{x}$) for each pedestrian based on the trajectories in the past $T$ time steps.
\subsubsection{Temporal Motion Tendency Learning}
We employ a recurrent neural network with one LSTM layer \cite{Hochreiter1997lstm} to capture the temporal motion dependency and predict future motion. Given the hidden state $\mathbf{h}_i^{t}$ of each pedestrian $i$ at time step $t$, a temporal extrapolation velocity $\bar{\mathbf{v}}_{i}^{t+1}$ can be obtained:
\begin{equation}
\mathbf{h}_i^{t} = \text{LSTM}(\mathbf{h}_i^{t-1}, \mathrm{Relu}(\phi(\mathbf{x}_i^t, \mathbf{v}_i^t))), \bar{\mathbf{v}}_{i}^{t+1} = \phi(\mathbf{h}_i^{t}),
\end{equation}
where $\phi(\cdot)$ represents Linear transformation, $\mathbf{x}_i^t$ and $\mathbf{v}_i^t$ are the current location and velocity.

\begin{figure}[t]
    \centering
    \includegraphics[width=.35\linewidth]{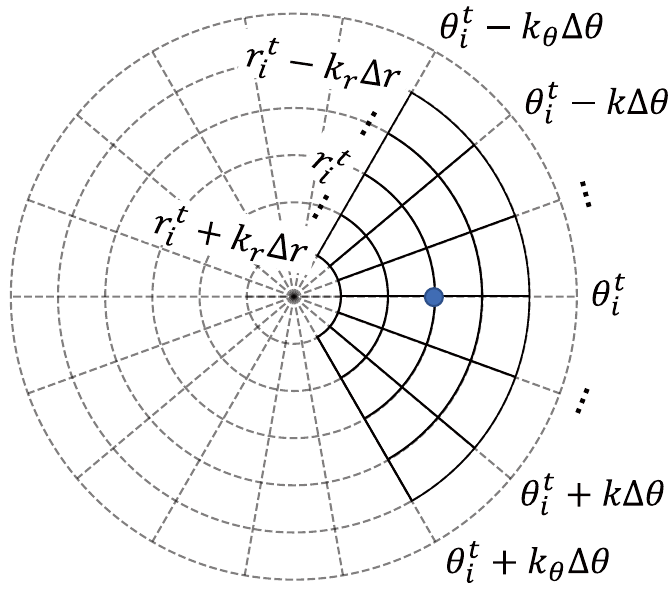}
    \caption{An illustration of the discrete candidate velocity space $V_{i}^t$ in a polar coordination system.
    The blue point represents the time extrapolation velocity $\bar{\mathbf{v}}_{i}^{t+1}$, with $r_i^t$ and $\theta_i^t$ denoting the magnitude and angle of $\bar{\mathbf{v}}_{i}^{t+1}$. 
    The polar space is discretized into a grid, with a predefined cell side length $\Delta r$ and $\Delta \theta$ for the magnitude and angle axes.
    The velocity candidates in $V_{i}^t$ are represented by the intersection points of the solid lines, centered at $\bar{\mathbf{v}}_{i}^{t+1}$ within $k_r$ grid cells on the magnitude axis and $k_\theta$ on the angle axis.}
    \label{fig:vspace}
\end{figure}

As human behavior is diverse and uncertain, multiple reasonable motion decisions exist for pedestrians. In our method, the possible motion decisions are explicitly modeled as the velocity candidates in a discrete candidate velocity space ${V}_{i}^{t}$, which is generated based on the temporal extrapolation velocity $\bar{\mathbf{v}}_{i}^{t+1}$. 
An illustration of ${V}_{i}^{t}$ is shown in Fig. \ref{fig:vspace}. ${V}_{i}^{t}$ is a velocity set with size $k_C = (2k_r+1)(2k_\theta+1)$.

\subsubsection{Coarse Trajectory Prediction}
After obtaining the candidate velocity space representing multiple motion decisions, we need to optimize for the best one as the coarse motion tendency for the subsequent time step.
We adopt the attention mechanism \cite{vaswani2017attention} to score the relation between the pedestrian's trajectories {$X_{i}^t = \{\mathbf{x}_i^{t-T+1}, \dots, \mathbf{x}_i^{t}\} \in \mathbb{R}^{T \times 4}$} in the past $T$ time steps and the velocity candidates from ${V}_{i}^{t} \in \mathbb{R}^{k_C \times 2}$. The attention score matrix $\widetilde{A}_i^t$ is calculated as follows:
\begin{equation}
\centering
\begin{aligned}
    &F_{i,P}^t = \text{MLP}_1(X_{i}^t),~F_{i,C}^t = \text{MLP}_2(V_{i}^t),\\
    &\widetilde{A}_i^t = \text{Softmax} \left(\frac{(F_{i,P}^{t} W_P) {(F_{i, C}^{t}W_C)}^\top}
    {\sqrt{d_F}}\right),
\end{aligned}
    \label{eq:attn}
\end{equation}
where $W_P,W_C$ are learnable parameters, $\sqrt{d_F}$ is the scaled factor for ensuring numerical stability \cite{vaswani2017attention}.
Then, the coarse preferred velocity $\widetilde{\mathbf{v}}_i^{t+1}$ of pedestrian $i$ at the next time step can be calculated by aggregating the information from the velocity candidate set ${V}_{i}^{t}$ with the weight of attention scores from Eq. \ref{eq:attn} as:
\begin{equation}
\label{eq:preferredvel}
   \widetilde{\mathbf{v}}_i^{t+1} = \text{MLP}_3(F^t_{i,P} + \widetilde{A}_i^t F_{i,C}^t).
\end{equation}
Then, the coarse preferred position is calculated as $\widetilde{\mathbf{x}}_i^{t+1} = \mathbf{x}_i^{t} + \widetilde{\mathbf{v}}_i^{t+1} \Delta t$, where $\Delta t$ is the horizon of a time step.

\subsection{Energy-Based Interaction Anticipating}
\label{subsec:interactionenergy}
\begin{figure}[t]
  \centering
  \subfloat[]{
    \label{subfig:interaction_area}
      \includegraphics[width=0.4\linewidth]{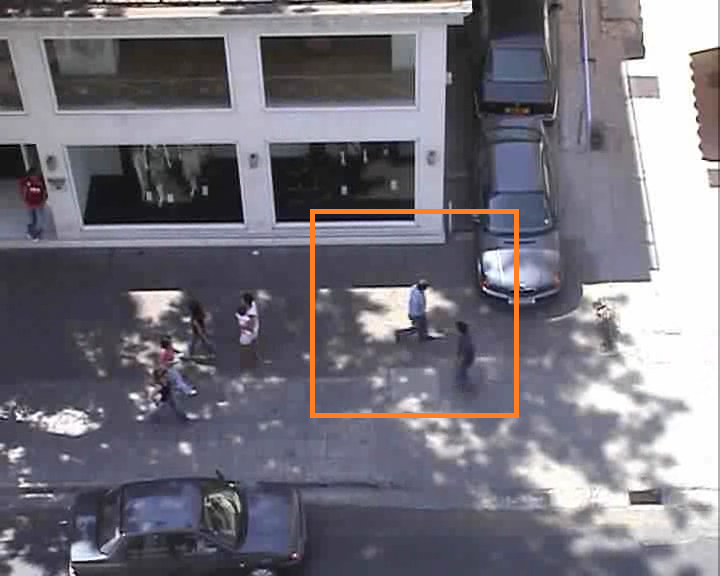}
      }
  \subfloat[]{
      \label{subfig:interaction_emap}
      \includegraphics[width=0.4\linewidth]{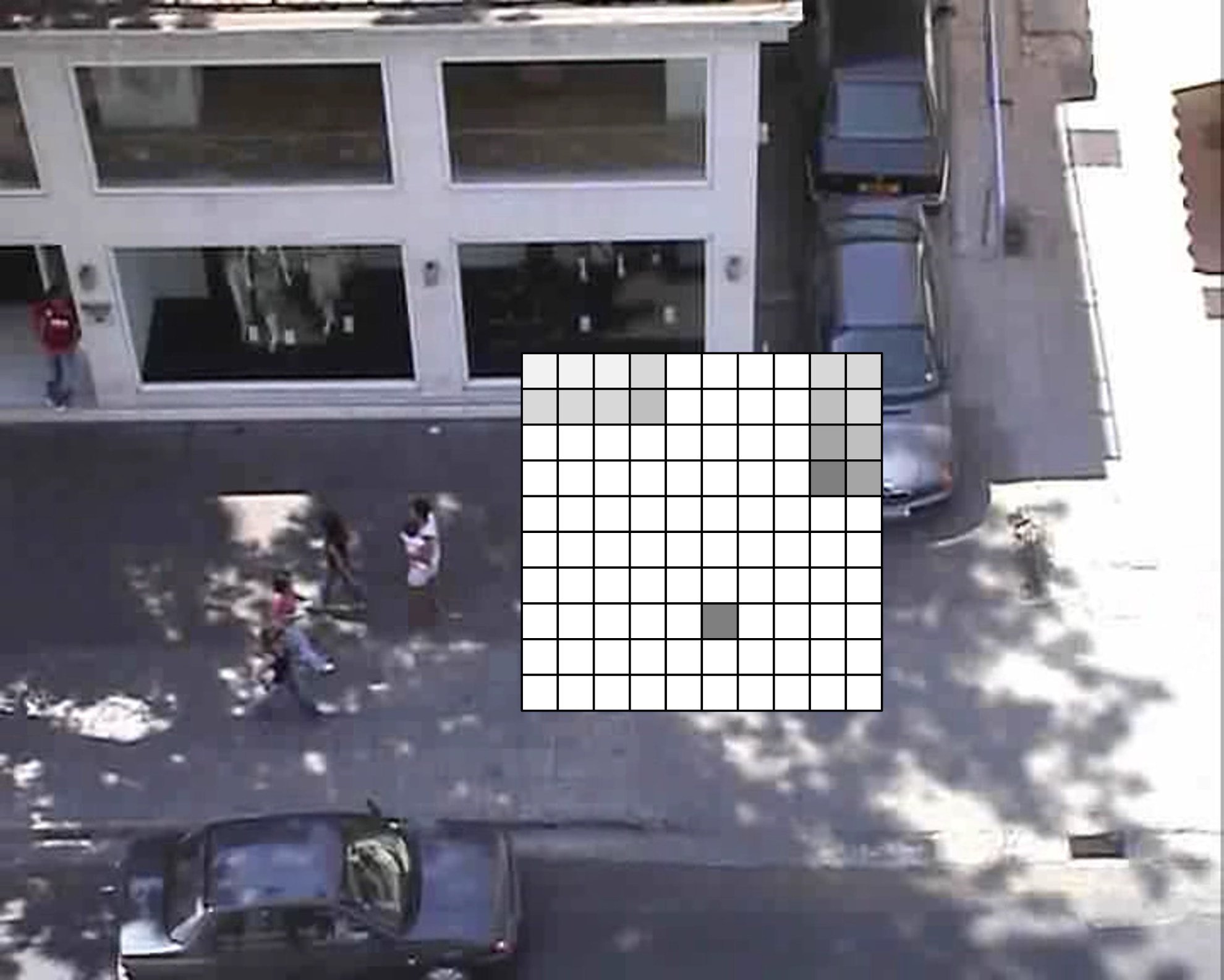}
      }
  \caption{An example of a pedestrian's square-shaped local interaction area. (a) The focal pedestrian at the center of the orange square interacts with heterogeneous neighbors within the square. (b) The interaction energy is recorded at the predicted local location of each neighbor (The darker color represents the higher value of interaction energy).}
  \label{fig:interaction_area}
\end{figure}

As humans anticipate and react to the future trajectories of their neighbors for collision avoidance, we employ an energy-based interaction model similar to \cite{xiang2023model} to calculate the interaction cost (i.e., repulsion intensity) driven by the coarse preferred velocity $\widetilde{\mathbf{v}}_i^{t+1}$ from Eq. \ref{eq:preferredvel}.
Our interaction model considers heterogeneous neighbors within a local neighborhood, which is a square area centering with the pedestrian (see Fig. \ref{subfig:interaction_area}), including pedestrians, static obstacles (e.g., buildings), and dynamic obstacles (e.g., vehicles).

\subsubsection{Interaction Energy}
Given the preferred velocity $\widetilde{\mathbf{v}}_i^{t+1}$ calculated by Eq. \ref{eq:preferredvel}, assuming that the neighbor $j$ holds its current velocity $\mathbf{v}_{j}^{t}$ for moving in the next time step, the predicted distance $\widetilde{d}_{ij}^{t+1}$ of pedestrian $i$ to the neighbor $j$ is calculated by considering the pedestrian $i$ may collide with their neighbor $j$ during a time step:
\begin{equation}
\label{equ:predict_dis}
    \widetilde{d}_{ij}^{t+1} = \Vert \mathbf{x}_{ij}^{t} + \widetilde{\mathbf{v}}_{ij}^{t+1} \cdot \widetilde{\tau}_{ij}^{t+1} \Vert_2,
\end{equation}
where $\mathbf{x}_{ij}^{t} = \mathbf{x}_{i}^{t}-\mathbf{x}_{j}^{t}$, $\widetilde{\mathbf{v}}_{ij}^{t+1} = \widetilde{\mathbf{v}}_{i}^{t+1} - \mathbf{v}_{j}^{t}$.
$\widetilde{\tau}_{ij}^{t+1} \in [0, \Delta t]$ is the predicted traveled time in the next time step, and it is obtained by solving the following quadratic function:
\begin{equation}
\label{eq:ttc}
\begin{aligned}
    &\tau_{ij}^{t+1} = {\arg\min}_{\tau} \Vert \mathbf{x}_{ij}^{t} + \widetilde{\mathbf{v}}_{ij}^{t+1} \cdot \tau \Vert_2, \\
    &\widetilde{\tau}_{ij}^{t+1} = \max(0,\min(\tau_{ij}^{t+1}, \Delta t)).
    \end{aligned}
\end{equation}

The interaction energy is calculated based on the predicted distance $\widetilde{d}_{ij}^{t+1}$:
\begin{equation}
\label{equ:energy}
    e_{ij}^{t} = e^{(1-\widetilde{d}_{ij}^{t+1}/d_{s})},
\end{equation}
where $d_{s}$ is a scaling factor. Higher interaction energy means that the pedestrian is more likely to collide with the neighbor and vice versa, which can be used to quantify the interaction cost of two objects.

\subsubsection{Socially Explainable Energy Map}
After calculating the interaction energies that anticipate and quantify the repulsion intensities from the neighbors,
we project the interaction energies onto an energy map $M_i^t$, which has the same size as the local interaction area, in order to explicitly indicate the socially anticipated occupancy of each point within the local interaction area.

$M_i^t$ is initialized as a zero matrix with size $L \times L$, where $L$ is the side length of the local interaction area. A zero value in $M_i^t$ means no occupancy, i.e., no risk of collision at this position during the next time step. The interaction energy $e_{ij}^{t}$ calculated by Eq. \ref{equ:energy} represents the occupancy of the neighbor $j$'s future location $\widetilde{\mathbf{x}}_j^{t+1} = \mathbf{x}_j^{t} + \mathbf{v}_j^t  \widetilde{\tau}_{ij}^{t+1}$. Fig. \ref{subfig:interaction_emap} illustrates the interaction energy map. Notably, to avoid performance degradation caused by the sparse matrix $M_i^{t}$, our method regards dynamic interaction neighbors as entities with a specific shape, and the points occupied inside the entity are allocated with the calculated interaction energy. A pedestrian neighbor is regarded as a square-shaped entity centered at $\widetilde{\mathbf{x}}_j^{t+1}$ with side length $\Delta L$; for other types of dynamic neighbors (e.g., vehicles and bicycles), as we don't specify the accurate type of those neighbors, their occupied points are those inside the bounding box from the raw dataset. For the static neighbors (e.g., buildings) from the scene segmentation, the interaction neighbors are the points labeled impassable to pedestrians.

For a point $\mathbf{p} = (p_x, p_y)$ in the local interaction area, the corresponding value in the energy map is calculated as:
\begin{equation}
    M_i^{t}[p_x, p_y] = \sum_{j \in \Omega(\mathbf{p})} w_{R(j)} \cdot e_{ij}^{t},
\end{equation}
where $\Omega(\mathbf{p})$ is the set of neighbors who are anticipated to occupy point $\mathbf{p}$, $R(j)$ is the type of neighbor $j \in \Omega(\mathbf{p})$, and $w_{R(j)}$ is the trainable weight of this neighbor type. The energy map contributes to the model for a better understanding of the future social relationship with the interaction neighbors.

\subsection{Multi-Modal Trajectory Prediction}
\label{subsec:cvae}

\begin{figure}[t]
    \centering
    \includegraphics[width=.8\linewidth]{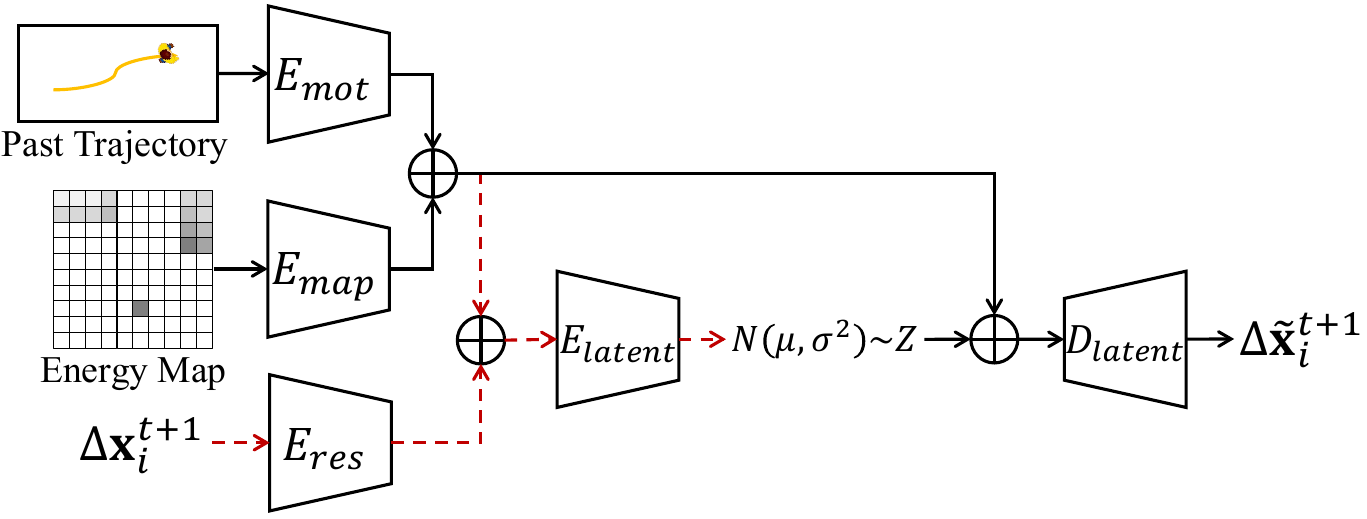}
    \caption{The architecture of the interaction-conditioned CVAE model. $\bigoplus$ represents the concatenation operation. Red dotted lines denote that the layers are only performed during training. All the components in the CVAE model are built using MLPs.}
    \label{fig:cvae}
\end{figure}
To capture the uncertainty of human motion, we employ an interaction-conditioned CVAE model for multimodal trajectory forecasting. The architecture of our CVAE model is illustrated in Fig. \ref{fig:cvae}. Different from the previous models that learn unexplainable randomness using CVAEs \cite{zhou2021scsr,yue2022nsp,zhou2023csr}, our model takes the pedestrian's past trajectories $X_{i}^{t}$ and the socially explainable energy map $M_i^t$ as input to reconstruct the position residual $\Delta \mathbf{x}_i^{t+1} = \mathbf{x}_i^{t+1} - \widetilde{\mathbf{x}}_i^{t+1}$ between the ground-truth future position $\mathbf{x}_i^{t+1}$ and the predicted coarse preferred position $\widetilde{\mathbf{x}}_i^{t+1}$. As a result, the CVAE can learn socially reasonable randomness from data.

In the training process, the CVAE model firstly obtains the encodings of motion $F_{i, M}^{t}$ and the encodings of the ground-truth position residual $F_{i, R}^{t+1}$:
\begin{equation}
\label{eq:cvae_input}
\begin{aligned}
    & F_{i,M}^{t} = E_\text{mot}(X_{i}^{t}) \oplus E_\text{map}(M_i^{t}),~
    F_{i,R}^{t+1} = E_\text{res}(\Delta \mathbf{x}_i^{t+1}), \\
\end{aligned}
\end{equation}
where $E_\text{mot}$, $E_\text{map}$ and $E_\text{res}$ are the encoders. Then, a latent encoder $E_\text{latent}$ generates the parameters $(\mathbf{\mu}_i^{t}, \mathbf{\sigma}_i^t)$ of a latent distribution:
\begin{equation}
  (\mathbf{\mu}_i^{t}, \mathbf{\sigma}_i^t) = E_\text{latent}(F_{i,M}^{t} \oplus F_{i,R}^{t+1}).
\end{equation}

When predicting the future trajectory of a pedestrian, a latent decoder $D_\text{latent}$ is employed to estimate a position residual $\Delta \widetilde{\textbf{x}}_i^{t+1}$:
\begin{equation}
    \Delta \widetilde{\textbf{x}}_i^{t+1} = D_\text{latent}(F_{i,M}^{t} \oplus Z_i^t),
\end{equation}
where the latent variable $Z_i^t$ is randomly sampled from a normal Gaussian distribution $\mathcal{N}(0,\mathbf{I})$. Finally, the predicted position is:
\begin{equation}
\label{equ:cvaeresult}
    \hat{\mathbf{x}}_i^{t+1} = \widetilde{\mathbf{x}}_i^{t+1} + \Delta \widetilde{\textbf{x}}_i^{t+1}.
\end{equation}

Notably, the CVAE aims to learn the social behaviors specifically referring to the subconscious steering/change of speed of pedestrians when they anticipate being close to their neighbors, which plays a crucial role in pedestrian motion \cite{kuang2008analysis,kuang2009subconscious}. To achieve this, we introduce collision anticipation/avoidance as a critical explainable social factor. This is because collision avoidance is the most notable social factor that contains rich social information, e.g., it can encode the social information that is intrinsically driven by the consideration of safety \cite{van2021algorithms}, as well as high-level factors such as culture, custom \cite{kaminka2018simulating}. Furthermore, collision is the only information that can be obtained reliably from trajectories. As the ground truth of other social factors (e.g., affective mind states, traffic signals, smartphone distraction, etc.) is unavailable, we do not explicitly model them as there is no way to quantitatively verify them. In addition, as energy-based models have been proven to effectively capture social interactions \cite{guy2010pledestrians, karamouzas2017implicit}, the interaction energy map is employed as the CVAE's condition, which enables the model to learn the subconscious behavior and explain its associated social interaction.

\subsection{Loss Function}
{Our model is trained end-by-end by minimizing a multi-task loss:
\begin{align}
\nonumber
    \mathcal{L} = \frac{1}{n M} \sum_{i=1}^{n} &\sum_{t=T+1}^{T+M}  \bigg(\lambda_{1} {\Vert \widetilde{\mathbf{x}}_i^{t}-\mathbf{x}_i^{t} \Vert_2} + \lambda_{2} {\Vert \Delta \widetilde{\mathbf{x}}_i^{t}-\Delta \mathbf{x}_i^{t} \Vert_2} \\
    &+ \lambda_{3} {D_\mathrm{KL}(\mathcal{N}(\mathbf{\mu}_i^{t}, \mathbf{\sigma}_i^t)||\mathcal{N}(0, \mathbf{I}))}  \bigg),
\end{align}
where $\lambda_{1}$, $\lambda_{2}$ and $\lambda_{3}$ are the loss weights. 
The first term is the position loss for training the coarse prediction model, which measures the distance between each preferred new position with the ground truth. The second term is the predicted position residual loss for training the CVAE model, which measures the distance between each predicted position residual with the ground truth. The third term is the Kullback-Leibler (KL) divergence loss for training the CVAE model, which measures the distance between the sampling distribution of the latent variable learned at the training stage with the sampling normal Gaussian distribution at the test stage.}

%% file: 3-Evaluation.tex
\section{Evaluation}
\subsection{Experiment Setup}
\subsubsection{Datasets}
To evaluate the effectiveness of our method, we conduct extensive experiments on two widely used datasets in pedestrian trajectory prediction tasks: ETH-UCY dataset \cite{pellegrini2009you, lerner2007crowds} and Stanford Drone Dataset (SDD) \cite{robicquet2016learning}. ETH-UCY includes pedestrians' trajectories in 5 scenes (ETH, HOTEL UNIV, ZARA1, and ZARA2). We follow the leave-one-out strategy \cite{mangalam2021ynet} for training and evaluation. SDD contains pedestrians' trajectories in 20 scenes. For SDD, we follow the data segmentation as \cite{yue2022nsp} for training and evaluation.
Following the common practice \cite{mangalam2021ynet,yue2022nsp}, the raw trajectories are segmented into 8-second trajectory segments with time step $\Delta t = 0.4s$, we train the model to predict the future 4.8$s$ (12 frames) based on the observed 3.2$s$ (8 frames).

\subsubsection{Evaluation Metrics}
We adopt the two widely used metrics, \textit{Average Displacement Error} (ADE) and \textit{Final Displacement Error} (FDE), to quantify the performance of our model. ADE computes the average $\ell_2$ distance between the prediction and the ground truth over all predicted time steps. FDE calculates the $\ell_2$ distance between the predicted final location and the ground-truth final location at the end of the prediction horizon. 
We follow the previously commonly used measurement to report the performances of the best of 20 predicted trajectories. Similar to \cite{zhou2021scsr,yue2022nsp, zhou2023csr}, we sample 20 future points at each prediction time step and select the best one as the predicted result.

\subsubsection{Environment}
Our model was implemented in PyTorch on a desktop computer running Ubuntu 20.04 containing an Intel $\circledR$ Core$^{\text{TM}}$ i7 CPU and an NVIDIA GTX 3090 GPU. The model is trained end-to-end with an Adam optimizer with a learning rate 0.0001. We trained the ETH-UCY for 100 epochs and SDD for 150 epochs. 

\subsection{Quantitative Evaluation}
\subsubsection{Quantitative Comparisons}
We compare \proj with state-of-the-art models in recent years. The experimental results on $\text{ADE}_{20}/\text{FDE}_{20}$ are presented in Tab. \ref{tab:compare_ethucy} for ETH-UCY and SDD, showing that \proj achieves state-of-the-art performance on both datasets. 
Compared with the SOTA baseline methods, our method achieves performance improvement by 66.67\% for FDE on ETH-UCY and 16.85\%/69.18\% for ADE/FDE on SDD. The main difference between \proj and the baseline methods is that our interaction-conditioned CVAE model learns socially reasonable motion randomness. The quantitative results demonstrate that \proj works well for better prediction performance. 
\begin{table*}[ht]
\centering
\begin{tabular}{cc|cccccc|c}
\hline
\multicolumn{2}{c|}{Model} & ETH       & Hotel      & UNIV      & ZARA1     & ZARA2     & AVG  & SDD \\ \hline
S-GAN             & \cite{gupta2018sgan} & 0.87/1.62 & 0.67/1.37  & 0.76/1.52 & 0.35/0.68 & 0.42/0.84 & 0.61/1.21 & 27.23/41.44\\
Sophie            & \cite{sadeghian2019sophie} & 0.70/1.43 & 0.76/1.67  & 0.54/1.24 & 0.30/0.63 & 0.38/0.78 & 0.51/1.15 & 16.27/29.38\\
Trajectron++      & \cite{salzmann2020trajectron++} & 0.39/0.83 & 0.12/0.21 & 0.20/0.44 & 0.15/0.33 & 0.11/0.25 & 0.19/0.41 & -\\
PECNet            & \cite{mangalam2020pecnet}  & 0.54/0.87 & 0.18/0.24 & 0.35/0.60 & 0.22/0.39 & 0.17/0.30 & 0.29/0.48 & 9.96/15.88\\
YNET              & \cite{mangalam2021ynet} & 0.28/0.33 & 0.10/0.16  & 0.24/0.41 & 0.17/0.27 & 0.13/0.22 & 0.18/0.27 & 7.85/11.85\\
Social-VAE        & {\cite{xu2022socialvae}} & 0.41/0.58 & 0.13/0.19  & 0.21/0.36 & 0.17/0.29 & 0.13/0.22 & -  & 8.10/11.72\\
CAGN              & \cite{duan2022cagn}& 0.41/0.65 & 0.13/0.23  & 0.32/0.54 & 0.21/0.38 & 0.16/0.33 & 0.25/0.43 & -\\
SIT               & \cite{shi2022sit} & 0.39/0.61 & 0.13/0.22  & 0.29/0.49 & 0.19/0.31 & 0.15/0.29 & 0.23/0.38 & 8.59/15.27\\
MUSE-VAE  & \cite{lee2022muse} & - & -  & - & - & - & -  & 6.36/11.10\\
MSRL              & \cite{wu2023msrl} & 0.28/0.47 & 0.14/0.22  & 0.24/0.43 & 0.17/0.30 & 0.14/0.23 & 0.19/0.33 & 8.22/13.39\\
S-CSR*            & \cite{zhou2021scsr}  & 0.19/0.35 & 0.06/0.07  & \underline{0.13}/0.21 & \underline{0.06}/0.07 & \underline{0.05}/0.08 & \underline{0.10}/0.16 & 2.77/3.45\\
NSP-SFM*          &   {\cite{yue2022nsp}}  & \underline{0.07}/\underline{0.09} & \underline{0.03}/\underline{0.07}  & \textbf{0.03}/\underline{0.04} & \textbf{0.02}/\underline{0.04} & \textbf{0.02}/\underline{0.04} & \textbf{0.03}/\underline{0.06} & \underline{1.78}/\underline{3.44}\\
CSR*              & \cite{zhou2023csr} & 0.28/0.53 & 0.07/0.08  & 0.24/0.35 & 0.07/0.09 & \underline{0.05}/0.09 & 0.14/0.23  & 4.87/6.32\\
Ours*             & -      & \textbf{0.06}/\textbf{0.04} & \textbf{0.025}/\textbf{0.01} & \textbf{0.03}/\textbf{0.03} & \textbf{0.02}/\textbf{0.01} & \textbf{0.02}/\textbf{0.01} & \textbf{0.03}/\textbf{0.02} & \textbf{1.48}/\textbf{1.06}\\ \hline
\end{tabular}
\caption{\label{tab:compare_ethucy} Quantitative comparison with state-of-the-art methods on ETH-UCY and SDD for $\text{ADE}_{20}/\text{FDE}_{20}$. The bold/underlined font represents the best/second best result. The prediction results on ETH-UCY and SDD are measured in meters and pixels, respectively. Previous SOTA methods labeled by * also employ the recursive prediction scheme.}
\end{table*}
\begin{table}[t]
\centering
\begin{tabular}{lccc}
\hline
\multicolumn{3}{c}{Components}                        & ADE/FDE                     \\ \cline{1-3}
$F_{goal}$ & Attention model & \begin{tabular}[c]{@{}c@{}}Interaction\\ -conditioned CVAE\end{tabular} &                             \\ \hline
-      & \checkmark    & \XSolidBrush               & 8.64/13.72                  \\
    -      & \XSolidBrush      & \checkmark             & 1.76/1.57                   \\
    
\checkmark & \XSolidBrush     & \checkmark               & 1.56/2.71 \\ 
    -      & \checkmark      & \checkmark               & \textbf{1.48}/\textbf{1.06} \\ \hline
\end{tabular}
\caption{\label{tab:ablation} Ablation study of different components of our method on the SDD dataset. $F_{goal}$ denotes the goal-attraction model proposed by NSP-SFM.}
\end{table}

\subsubsection{Ablation Study.}
We conduct ablative experiments to show the effectiveness of the key components in our model.

\textbf{Ablating the interaction-conditioned CVAE}. In this experiment (named Ours/wo), we connect the coarse prediction model in \proj with the same CVAE model as \cite{zhou2021scsr,yue2022nsp,zhou2023csr}, which is only conditioned on the past trajectories, to learn the random residuals for the predicted preferred position from the coarse prediction model. Tab. \ref{tab:ablation} shows the quantitative results on SDD. Because the model doesn't consider pedestrian interactions and learns unexplainable motion randomness, compared with our full model, significant performance degradation occurs on both ADE and FDE, demonstrating the importance of our interaction-conditioned CVAE model for achieving better performance.

\textbf{Ablating the attention model}. In this experiment, we use the temporal extrapolation velocity generated by the temporal encoder as the output coarse preferred velocity of the coarse prediction model. The prediction results on SDD in Tab. \ref{tab:ablation} show performance degradation compared to our full model. However, when compared with the SOTA baselines, it still achieves better prediction accuracy, demonstrating the importance of our proposed interaction-conditioned CVAE model which learns the uncertainty of human motions.

\textbf{Ablating the coarse prediction model}. We also conduct another ablation experiment, named G\proj, by replacing the coarse motion prediction model in Sec. \ref{subsec:coarseprediction} with the goal-attraction model from the SOTA NSP-SFM method \cite{yue2022nsp}, to further demonstrate the importance of the interaction-conditioned multimodal learning scheme employed in \proj. Tab. \ref{tab:ablation} gives the quantitative results of G\proj on SDD, showing performance degradation compared with our full model. 
However, when compared with NSP-SFM, G\proj achieves better performance with 11.80\% improvement on ADE and 20.35\% improvement on FDE, demonstrating the better prediction capability of our energy interaction-conditioned CVAE model for human motion uncertainty learning.

\begin{figure}[t]
\centering
\begin{minipage}{\linewidth}
\centering
	\includegraphics[width=0.313\linewidth]{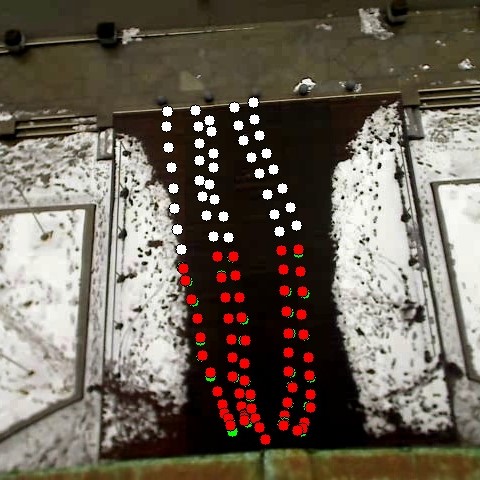}	
	\includegraphics[width=0.313\linewidth]{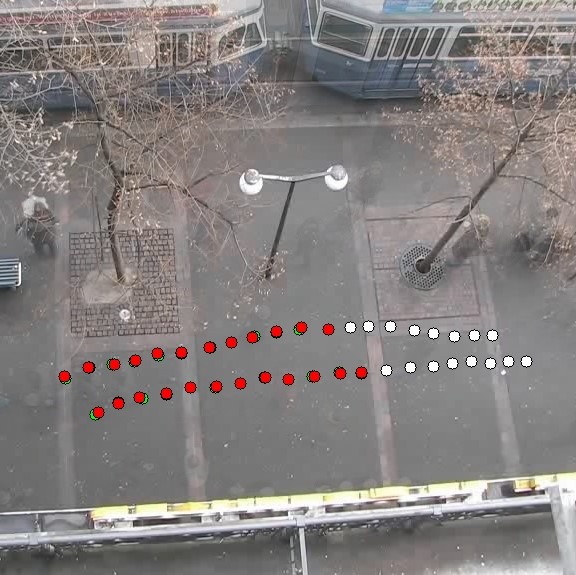}
	\includegraphics[width=0.313\linewidth]{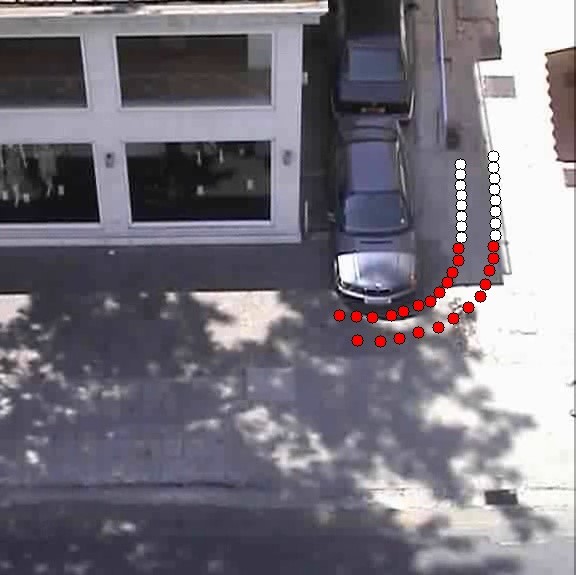}
\end{minipage}
\begin{minipage}{\linewidth}
\centering
	\includegraphics[width=0.313\linewidth]{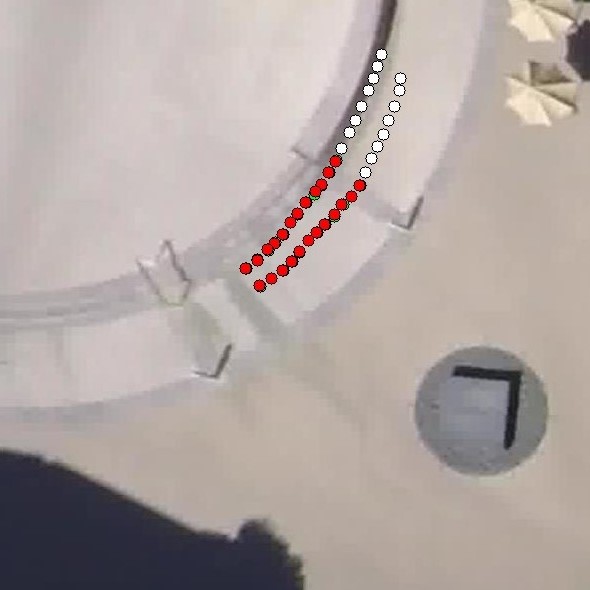}
	\includegraphics[width=0.313\linewidth]{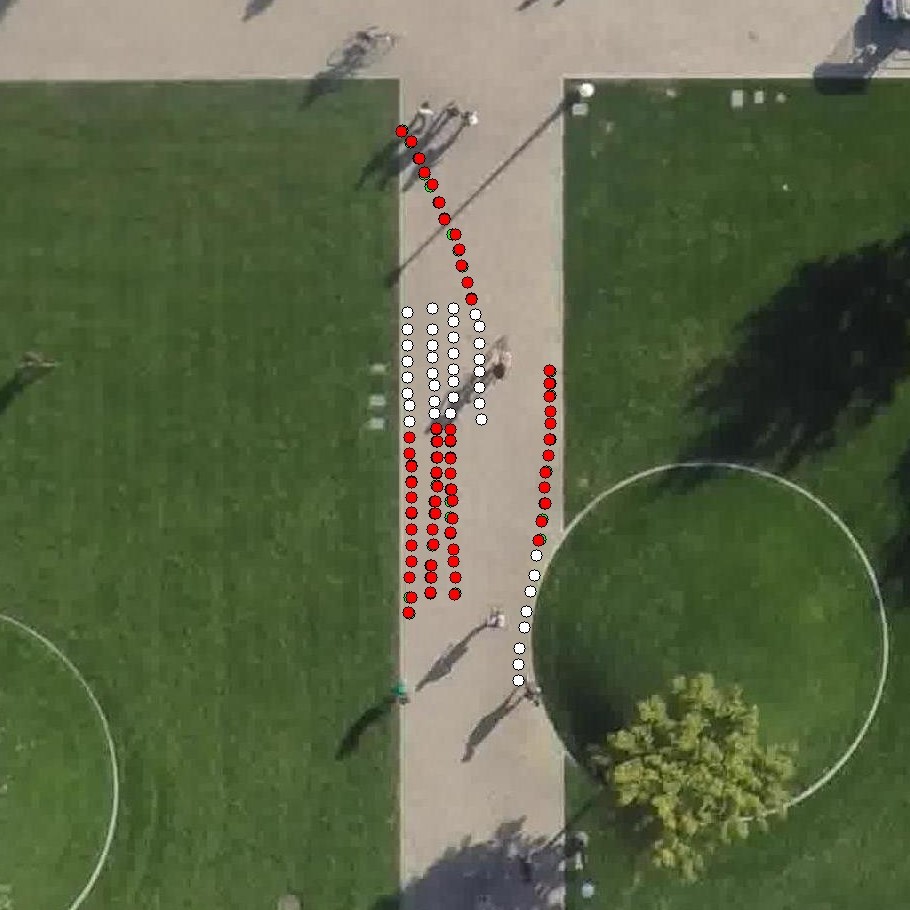}
	\includegraphics[width=0.313\linewidth]{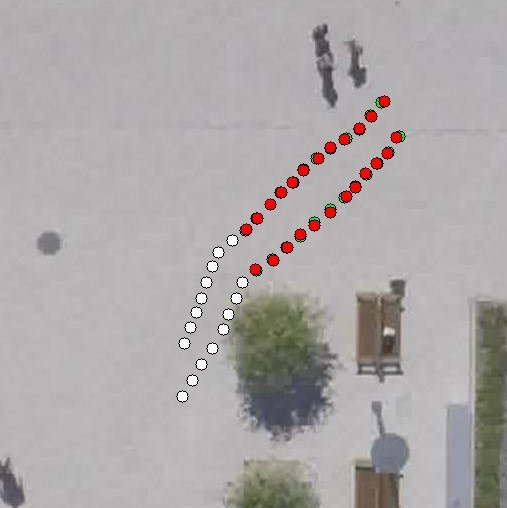}
\end{minipage}
	
 \caption{Visualization results of our method. The visualized trajectories are the best predictions sampled from 20 trials. The white, green, and red dots represent the observed, ground-truth, and predicted trajectories respectively.}
 \label{fig:visual_result}
\end{figure}
\begin{figure}[t]
\centering
\begin{minipage}{\linewidth}
\centering
	\subfloat[NSP-SFM-1]{
		\label{subfig:viscompare_sota_1}
		\includegraphics[width=0.313\linewidth]{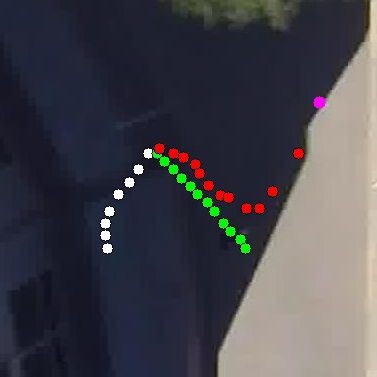}	
	}
	\subfloat[NSP-SFM-2]{
		\label{subfig:viscompare_sota_2}
		\includegraphics[width=0.313\linewidth]{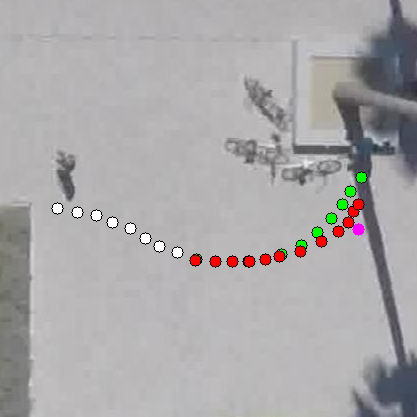}
	}
        \subfloat[NSP-SFM-3]{
		\label{subfig:viscompare_sota_3}
		\includegraphics[width=0.313\linewidth]{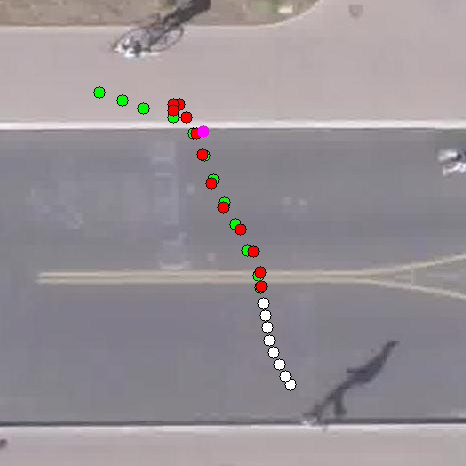}
	}
\end{minipage}
\begin{minipage}{\linewidth}
\centering
	\subfloat[G\proj-1]{
		\label{subfig:viscompare_gcvae_1}
		\includegraphics[width=0.313\linewidth]{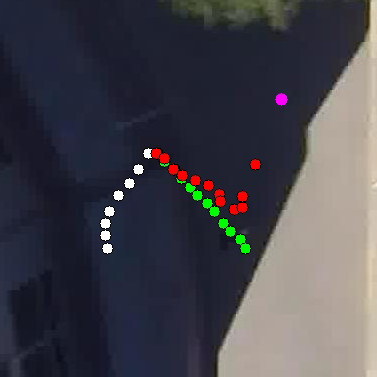}	
	}\noindent
	\subfloat[G\proj-2]{
		\label{subfig:viscompare_gcvae_2}
		\includegraphics[width=0.313\linewidth]{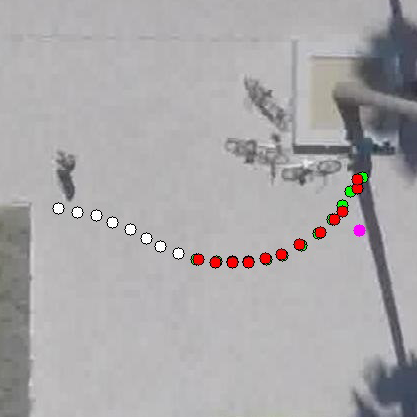}
	}
        \subfloat[G\proj-3]{
		\label{subfig:viscompare_gcvae_3}
		\includegraphics[width=0.313\linewidth]{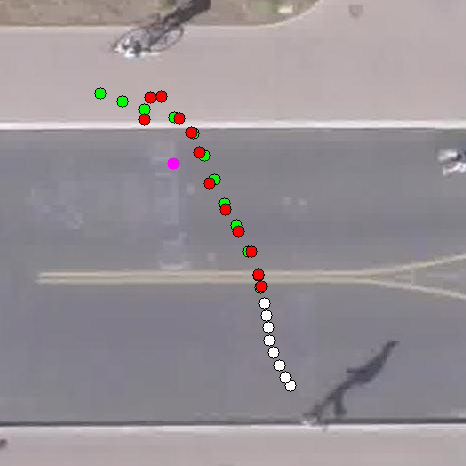}
	}
\end{minipage}
\begin{minipage}{\linewidth}
\centering
	\subfloat[Ours-1]{
            \label{subfig:viscompare_our_1}
		\includegraphics[width=0.313\linewidth]{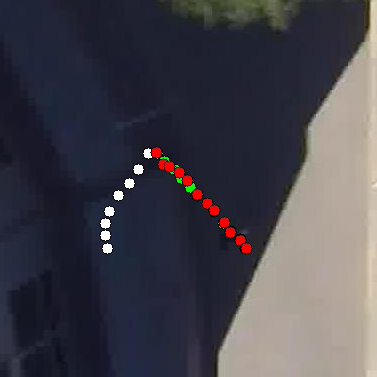}
	}
	\subfloat[Ours-2]{
		\label{subfig:viscompare_our_2}
		\includegraphics[width=0.313\linewidth]{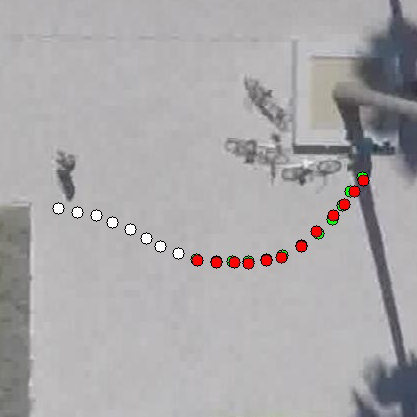}
	}
	\subfloat[Ours-3]{
		\label{subfig:viscompare_our_3}
		\includegraphics[width=0.313\linewidth]{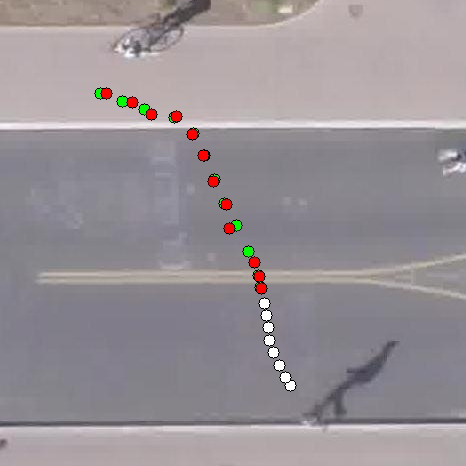}
	}
\end{minipage}
\caption{Visualization comparisons with NSP-SFM and G\proj. The visualized trajectories are the best predictions sampled from 20 trials.
Our method predicts future trajectories closer to the ground truth than compared methods.
The purple red dots in the visualization results of NSP-SFM and G\proj represent the sampled goals.}
\label{fig:visual_compare}
\end{figure}
\begin{figure}[t]
\centering
\begin{minipage}{\linewidth}
\centering
	\subfloat[NSP-SFM-1]{
		\label{subfig:multimoda_nsp_1}
		\includegraphics[width=0.313\linewidth]{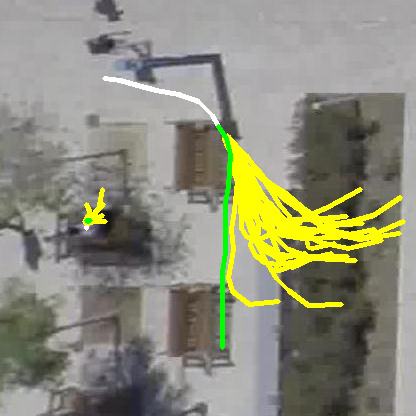}	
	}
	\subfloat[NSP-SFM-2]{
		\label{subfig:multimoda_nsp_2}
		\includegraphics[width=0.313\linewidth]{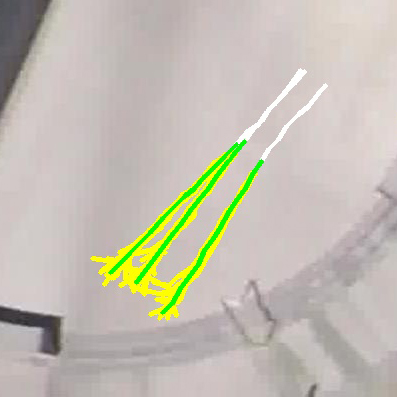}
	}
        \subfloat[NSP-SFM-3]{
		\label{subfig:multimoda_nsp_4}
		\includegraphics[width=0.313\linewidth]{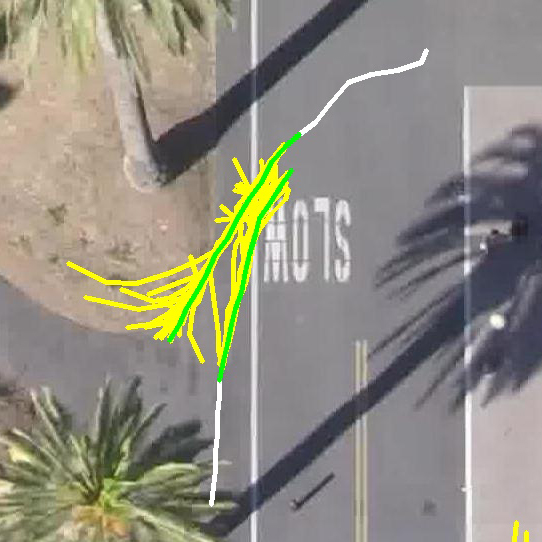}
	}
\end{minipage}
\begin{minipage}{\linewidth}
\centering
	\subfloat[Ours/wo-1]{
		\label{subfig:multimoda_ab_1}
		\includegraphics[width=0.313\linewidth]{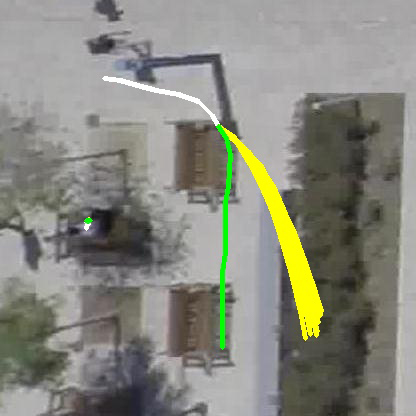}	
	}
	\subfloat[Ours/wo-2]{
		\label{subfig:multimoda_ab_2}
		\includegraphics[width=0.313\linewidth]{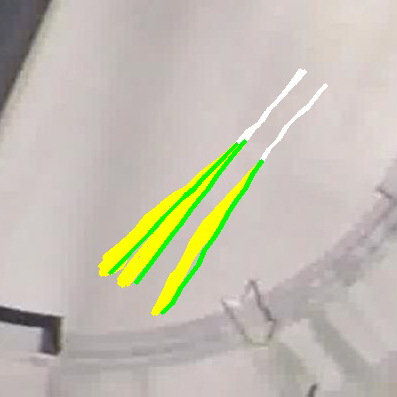}
	}
        \subfloat[Ours/wo-3]{
		\label{subfig:multimoda_ab_4}
		\includegraphics[width=0.313\linewidth]{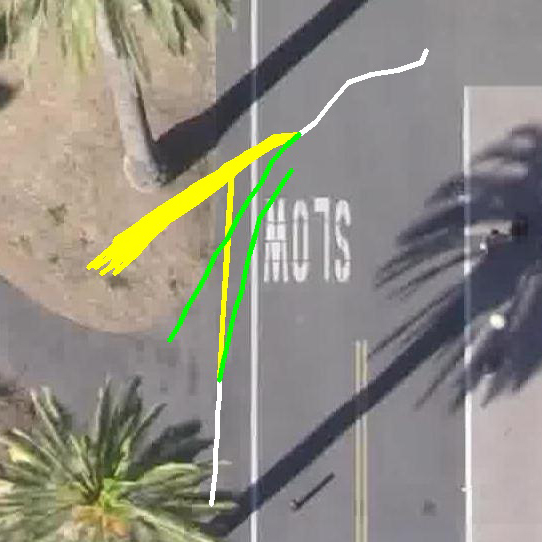}
	}
\end{minipage}
\begin{minipage}{\linewidth}
\centering
	\subfloat[Ours-1]{
		\label{subfig:multimoda_our_1}
		\includegraphics[width=0.313\linewidth]{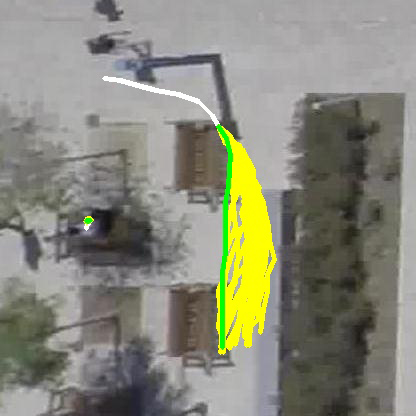}	
	}
	\subfloat[Ours-2]{
		\label{subfig:multimoda_our_2}
		\includegraphics[width=0.313\linewidth]{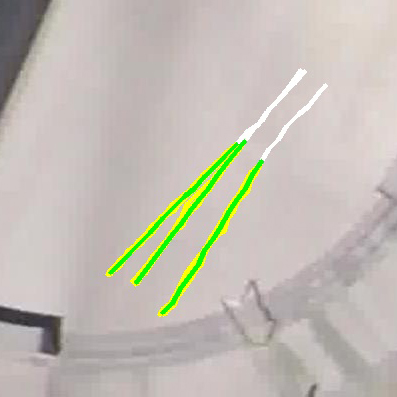}
	}
        \subfloat[Ours-3]{
		\label{subfig:multimoda_our_4}
		\includegraphics[width=0.313\linewidth]{figures/multicompare/our_391_little_video3.jpg}
	}
\end{minipage}
\caption{Visualization comparisons of the multiple predicted trajectories with NSP-SFM and \proj without the interaction-conditioned CVAE (Ours/wo). Our method predicts more socially reasonable future trajectories than the compared methods.
The white and green lines are the observed and ground-truth trajectories. The yellow lines for each pedestrian are the 20 predicted trajectories.}
\label{fig:visual_multimodal}
\end{figure}

\subsection{Qualitative Evaluation}
We first visualize the predicted trajectories in several scenarios to illustrate the effectiveness of our method. The visualization results are shown in Fig. \ref{fig:visual_result}. 

\textbf{Predicted trajectory comparison.} To further validate the better performance of our model, in Fig. \ref{fig:visual_compare}, we compare our visualization results with the SOTA NSP-SFM model \cite{yue2022nsp}. NSP-SFM may predict trajectories that obviously deviate from the ground-truth final positions. This is because NSP-SFM learns force-driven steering behaviors plus with unexplainable motion randomness; the predicted results show strong determinism in reaching a sampled final goal. In contrast, \proj employs an energy interaction-conditioned CVAE model for learning socially reasonable human motion uncertainty, thus achieving better prediction performance. 

In Fig. \ref{fig:visual_compare}, we also compare with the visualization results of the aforementioned ablation model G\proj. Due to the determinism nature of the goal-attraction model \cite{yue2022nsp}, compared with our full model's result (Figs. \ref{subfig:viscompare_our_1}-\ref{subfig:viscompare_our_3}), the predicted trajectories of G\proj show slight deviation from the ground-truth trajectories because the predicted goal is far from the ground truth. However, G\proj shows better visual results than NSP-SFM, demonstrating the proposed method's prediction capability to achieve better performance.

\textbf{Interaction-conditioned multimodal prediction.}
As shown in Fig. \ref{fig:visual_multimodal}, we compare the multiple predicted trajectories of the NSP-SFM, the ablation experiment of \proj without the interaction-conditioned CVAE (Ours/wo), and our full model (Ours). Our full model's results in Figs. \ref{subfig:multimoda_our_1}-\ref{subfig:multimoda_our_4} demonstrate that by conditioning on the socially explainable interaction energy map, \proj learns better human motion uncertainty than the model without conditioned on interaction. Figs. \ref{subfig:multimoda_our_2} and \ref{subfig:multimoda_our_4} also demonstrate that \proj can predict socially reasonable trajectories for avoiding potential collisions than the models without conditioned on interaction.

%% file: 4-Conclusion.tex
\section{Conclusion}
In this work, we present \proj, a novel multimodal pedestrian trajectory prediction method with an interaction-conditioned CVAE model for learning socially reasonable human motion randomness. \proj explicitly models the anticipated social relationships of pedestrians and their neighbors by using an interaction energy map generated based on an energy-based interaction model. Taking the interaction energy map as a condition, the CVAE model can learn the uncertainty of human motions while maintaining social awareness. The proposed method outperforms existing state-of-the-art methods in achieving higher prediction accuracy. One limitation is that our method is computationally inefficient as we sequentially predict the energy map for each pedestrian. In the future, we will improve the computation performance by exploring other formulations of energy-based interaction.

%% file: 6-acknowledgement.tex
\clearpage
\section*{Acknowledgments} Xiaogang Jin was supported by the National Natural Science Foundation of China (Grant No. 62036010). He Wang has received funding from the European Union’s Horizon 2020 research and innovation programme under grant agreement No 899739 CrowdDNA.

%% file: 5-Appendix.tex

\part*{Appendix}
\appendix
\section{Energy-based Interaction Anticipating}
\paragraph{Time to Collision.}
We provide a detailed explanation of how to solve the quadratic function $\tau_{ij}^{t+1} = {\arg\min}_{\tau} \Vert \mathbf{x}_{ij}^{t} + \widetilde{\mathbf{v}}_{ij}^{t+1} \cdot \tau \Vert_2$ to obtain $\tau_{ij}^{t+1}$, which denotes the time taken for pedestrian $i$ to collide with a neighbor $j$ (time to collision, i.e., when the predicted distance is 0).
We predict collisions by assuming a linear extrapolation of the positions of the pedestrian and the neighbor based on the pedestrian's coarse preferred velocity and the neighbor's current velocity. Then determine whether there is a time $\tau^*$ at which the two future linear trajectories intersect, i.e.,
\begin{equation}
\centering
\label{equ:quadratic_func}
     \Vert \mathbf{x}_{ij}^{t} + \widetilde{\mathbf{v}}_{ij}^{t+1} \cdot \tau^* \Vert_2 = 0,
\end{equation}
by rearranging, we obtain the solution
\begin{equation}
\centering
\label{equ:sovler}
\tau^* =  -\frac{\Vert \mathbf{x}_{ij}^{t} \Vert^2}{\mathbf{x}_{ij}^{t} \cdot \widetilde{\mathbf{v}}_{ij}^{t+1}}.
\end{equation}
If the solution $\tau^*$ doesn't exist (i.e., $\mathbf{x}_{ij}^{t} \cdot \widetilde{\mathbf{v}}_{ij}^{t+1} = 0$ ) or is negative, $\tau_{ij} = \infty$, which means the pedestrian and the neighbor will not collide in the future. Otherwise, $\tau_{ij} = \tau^*$.

\section{Evaluation}
\subsection{Details of Datasets}
\textbf{The ETH-UCY dataset} \cite{pellegrini2009you, lerner2007crowds} includes pedestrians' trajectories in 5 scenes (ETH, HOTEL UNIV, ZARA1, and ZARA2), including more than 1500 pedestrians and thousands of nonlinear trajectories with various interactions. The trajectories are recorded in the world coordinates (i.e., meters).
We consider the pedestrian-pedestrian, pedestrian-static obstacle (e.g., buildings) interactions for ETH-UCY. 
We follow the leave-one-out strategy \cite{mangalam2021ynet,yue2022nsp} for training and evaluation, i.e.,  training our model on four sub-datasets and testing it on the remaining one. Following the common practice \cite{mangalam2021ynet,yue2022nsp}, the raw trajectories are segmented into 8-second trajectory segments with time step $\Delta t = 0.4$ seconds, we train the model to predict the future 4.8 seconds (12 frames) based on the observed 3.2 seconds (8 frames). Since our model works in the image coordinate space (i.e., pixel space), we project the world coordinates in ETH-UCY into the pixel using the homography matrices provided in Y-net \cite{mangalam2021ynet} for training and testing, and then project the pixel results to meters for calculating quantitative metrics.

\textbf{The SDD dataset} \cite{robicquet2016learning} contains videos of a university campus with six classes of traffic agents with rich interactions, including about 185,000 interactions between different agents
and approximately 40,000 interactions between the agent and the environment. The trajectories are recorded in pixels.
More complex interactions are considered in SDD, including pedestrian-pedestrian, pedestrian-static obstacle (e.g., buildings), and pedestrian-dynamic obstacle (e.g., vehicles, bicycles) interactions.
We extract the trajectories like that of ETH-UCY to train the model for predicting the future 4.8$s$ (12 frames) based on the observed 3.2$s$ (8 frames).


\subsection{Qualitative Comparisons}
\subsubsection{Predicted trajectories} 
\begin{figure*}[t]
    \centering
    \begin{minipage}{\linewidth}
    \centering
    \includegraphics[width=.16\linewidth]{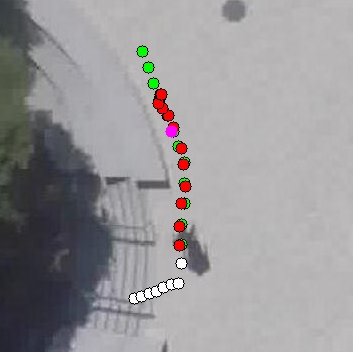}
    \includegraphics[width=.16\linewidth]{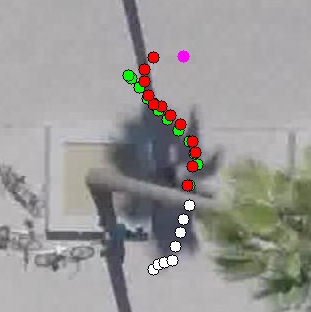}
    \includegraphics[width=.16\linewidth]{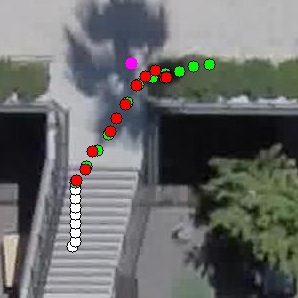}
    \includegraphics[width=.16\linewidth]{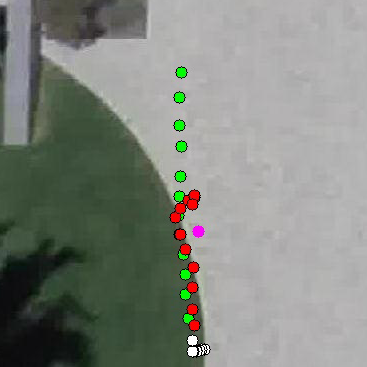}
    \includegraphics[width=.16\linewidth]{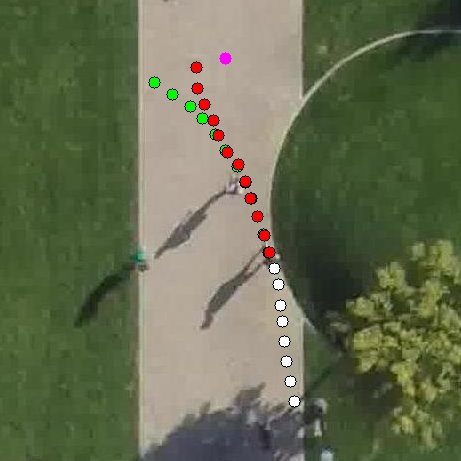}
    \includegraphics[width=.16\linewidth]{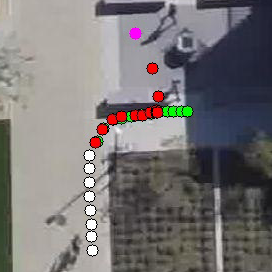}
    \end{minipage}
    \begin{minipage}{\linewidth}
    \centering
    \includegraphics[width=.16\linewidth]{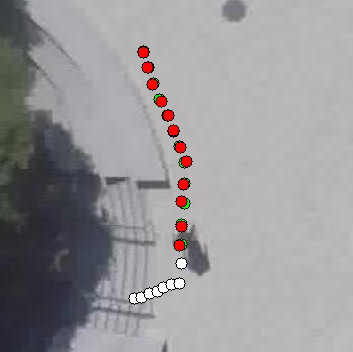}
    \includegraphics[width=.16\linewidth]{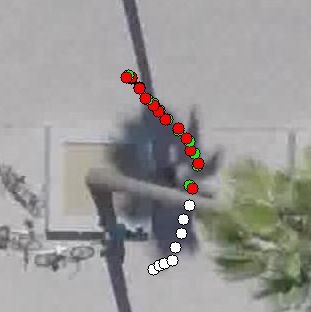}
    \includegraphics[width=.16\linewidth]{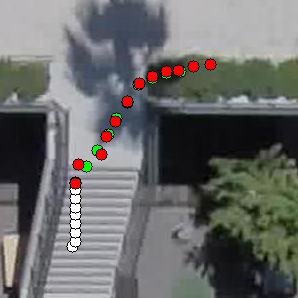}
    \includegraphics[width=.16\linewidth]{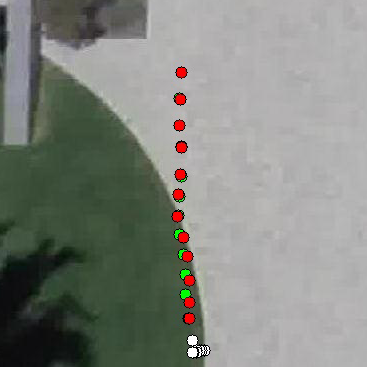}
    \includegraphics[width=.16\linewidth]{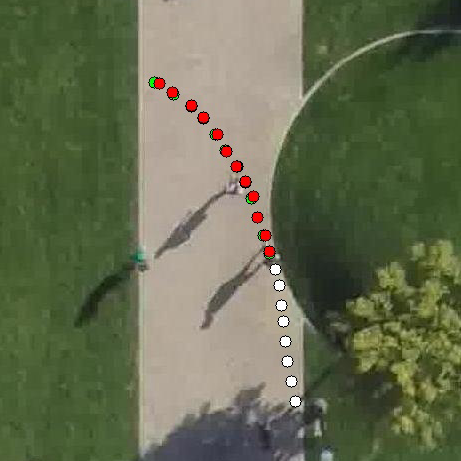}
    \includegraphics[width=.16\linewidth]{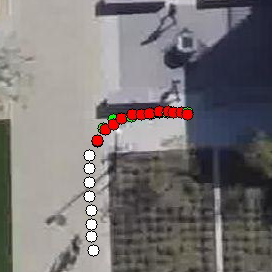}
    \end{minipage}
    \caption{Supplementary materials for visualization comparisons with NSP-SFM. The results of NSP-SFM and our method are shown in the first and second rows, respectively. The visualized trajectories are the best predictions sampled from 20 trials. The white, green, and red dots represent the observed, ground-truth, and predicted trajectories respectively. The purple red dots in the visualization results of NSP-SFM (i.e., the results in the first row) represent the sampled goals.}
    \label{fig:visualcomparison_prediction}
\end{figure*}

As shown in Fig. \ref{fig:visualcomparison_prediction}, we provide more visualization results of the predicted trajectories and compare them with that of the state-of-the-art method NSP-SFM \cite{yue2022nsp}. Results show that our method predicts future trajectories closer to the ground truth than NSP-SFM.
The predicted results of NSP-SFM (i.e., results exhibited in the first row) show strong determinism in reaching a sampled final goal driven by a goal-attraction model, resulting in the predicted trajectories deviating from the ground-truth final positions. In contrast, our method employs an interaction-conditioned CVAE model for learning socially reasonable human motion uncertainty, thus achieving better prediction performance.

\subsubsection{Multimodal prediction}
\begin{figure*}[t]
    \centering
    \begin{minipage}{\linewidth}
    \centering
    \includegraphics[width=.16\linewidth]{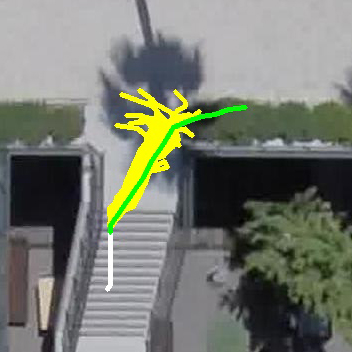}
    \includegraphics[width=.16\linewidth]{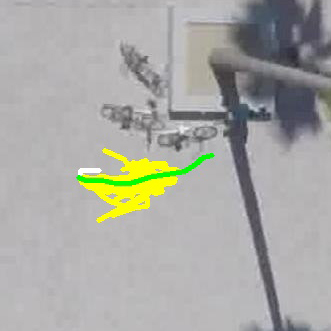}
    \includegraphics[width=.16\linewidth]{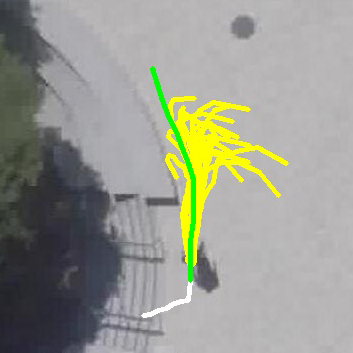}
    \includegraphics[width=.16\linewidth]{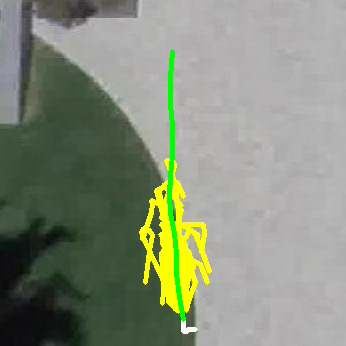}
    \includegraphics[width=.16\linewidth]{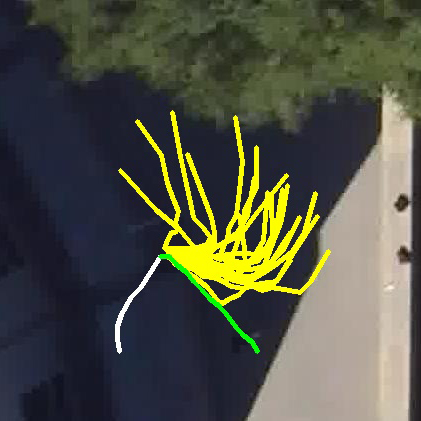}
    \includegraphics[width=.16\linewidth]{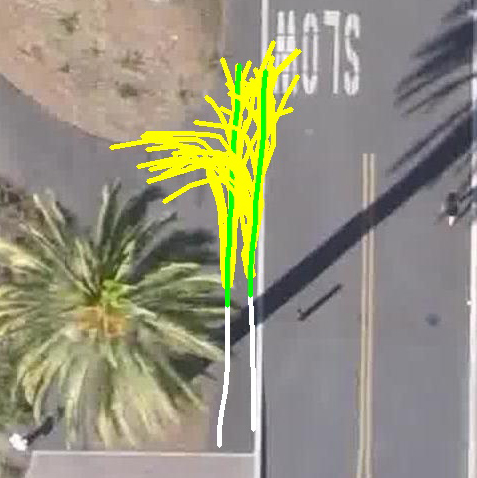}
    \end{minipage}
    \begin{minipage}{\linewidth}
    \centering
    \includegraphics[width=.16\linewidth]{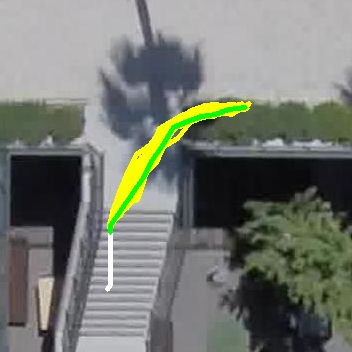}
    \includegraphics[width=.16\linewidth]{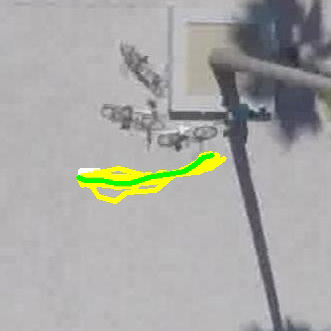}
    \includegraphics[width=.16\linewidth]{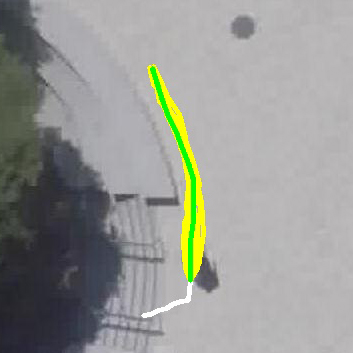}
    \includegraphics[width=.16\linewidth]{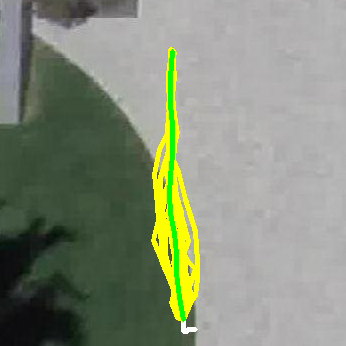}
    \includegraphics[width=.16\linewidth]{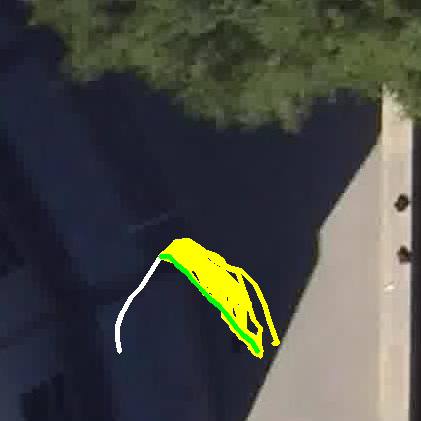}
    \includegraphics[width=.16\linewidth]{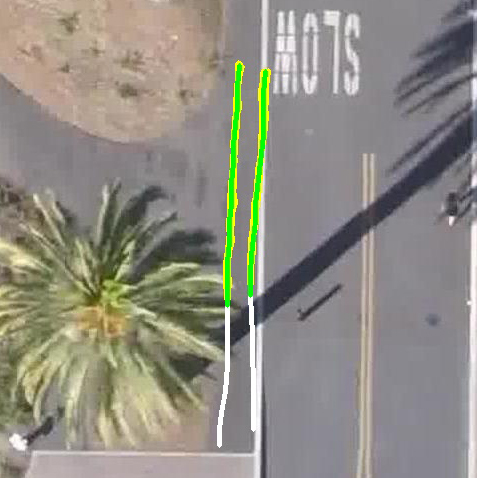}
    \end{minipage}
    \caption{Supplementary materials for visualization comparisons of multiple predicted trajectories with NSP-SFM. The results of NSP-SFM and our method are shown in the first and second rows, respectively. The white and green lines are the observed and ground-truth trajectories. The yellow lines for each pedestrian are the 20 predicted trajectories.}
    \label{fig:visualcomparison_multimoda}
\end{figure*}

As shown in Fig. \ref{fig:visualcomparison_multimoda}, we provide more visualization results of the multiple predicted trajectories and compare them with that of the state-of-the-art method NSP-SFM \cite{yue2022nsp}. The better results of our method exhibited in the second row demonstrate that, by conditioning on the socially explainable interaction energy map, our method learns better human motion uncertainty than NSP-SFM which doesn't consider social interactions in human motion randomness learning.

\subsection{Quantitative Comparisons}
The quantitative results of our model show a smaller number of FDE than ADE in some test cases. Actually, it is not rare that FDE is superior to ADE for trajectory prediction, similar phenomena can be observed from both NSP-SFM (one scene, Tab. 1) and its follow-up BNSP-SFM \cite{yue2023human} (systematically, Tab. 4). Both BNSP-SFM and our model use the recursive prediction scheme to gradually update future trajectories, with the $T$-step observed and previously predicted trajectories as input for each prediction. We train the model by minimizing the prediction error at each time step, and the closer the input past trajectories are to the real one, the more accurate our model’s predictions will be.

\subsection{Experiment Setup}
\subsubsection{Hyperparamters}
The hyperparameters utilized in our experiments are shown in Tab. \ref{tab:hyperpara}. The same set of hyperparameters was employed for both the ETH-UCY and SDD datasets.
\begin{table}[t]
\centering
\begin{tabular}{ccccc}
\hline
$\Delta \theta$   & $k_\theta$   & \begin{tabular}[c]{@{}c@{}}$\Delta r$ \\ (pixel)\end{tabular} & $k_r$   & \begin{tabular}[c]{@{}c@{}}$d_s$\\ (pixel) \end{tabular} \\ \hline
$\frac{1}{6}\pi$   &  4     &   5   &    2   &  5    \\ \hline \hline
\begin{tabular}[c]{@{}c@{}}$L$ \\ (pixel)\end{tabular} & \begin{tabular}[c]{@{}c@{}}$\Delta L$\\ (pixel) \end{tabular} & $\lambda_1$  & $\lambda_2$ & $\lambda_3$   \\ \hline
100     &     5 &     1 &   0.5 &  0.5     \\ \hline
\end{tabular}
\caption{\label{tab:hyperpara} The hyperparameters utilized for implementing SocialCVAE.}

\end{table}

\subsubsection{Learnable parameters}
We also show the learned weights of different interaction energies for ETH-UCY and SDD in Tab. \ref{tab:learnablePara}. It is worth noting that all learnable weights were initialized as 1 prior to training the model.
For simplicity, the weight of the interaction energy with pedestrian neighbors was not trained and remained at the value 1 in all experiments.
\begin{table}[h]
\centering
\begin{tabular}{c|c|c}
\hline
Learnable Parameter & $w_{s}$ & $w_{d}$ \\ \hline
ETH                 & 1.0229   &  -       \\ 
HOTEL               & 1.0081   &  -       \\ 
UNIV                & 1.2104   &  -       \\ 
ZARA1               & 1.0497   &  -       \\ 
ZARA2               & 1.1429   &  -       \\ 
SDD                 & 1.0644   &  1.1887      \\ \hline
\end{tabular}
\caption{\label{tab:learnablePara}The learned weights of different interaction energies. $w_{s}$ and $w_{d}$ respectively represent the learned weight of interaction energy with the static and dynamic obstacles.
}
\end{table}

\subsubsection{Sub-network architecture}
We also provide the detailed network architectures of the sub-networks employed in our experiments in Tab. \ref{tab:subnectworks}. The ReLU activation function is used in our network for the non-linearity of the network. The network configurations for both ETH-UCY and SDD datasets were identical.
\begin{table}[h]
\centering
\footnotesize
\begin{tabular}{cc|c}
\hline
\multicolumn{2}{c|}{Sub-network}                                                                                          & Network Architecture                               \\ \hline
\multicolumn{1}{c|}{\multirow{5}{*}{Section 3.2}} & LSTM                                                                  & $[4, 256 ,64]$                             \\
\multicolumn{1}{c|}{}                             & $MLP_1$                                                               & $[32 , 1024 , 512 , 64]$                 \\
\multicolumn{1}{c|}{}                             & $MLP_2$                                                                   & $[2 , 1024 , 512 , 64]$                     \\
\multicolumn{1}{c|}{}                             & $MLP_3$                                                                   & $[64 , 1024 , 512 , 2]$                     \\
\multicolumn{1}{c|}{}                             & \begin{tabular}[c]{@{}c@{}}Linear\\ (query encoding)\end{tabular}      & $[64 , 64]$                                       \\
\multicolumn{1}{c|}{}                             & \begin{tabular}[c]{@{}c@{}}Linear\\ (key encoding)\end{tabular} & $[64 , 64]$                                       \\ \hline
\multicolumn{1}{c|}{\multirow{5}{*}{Section 3.4}} & $E_{mot}$                                                               & $[16 , 512 , 256 , 16]$                     \\
\multicolumn{1}{c|}{}                             & $E_{map}$                                                                 & $[10000 , 1024 , 512 , 256 , 32]$ \\
\multicolumn{1}{c|}{}                             & $E_{res}$                                                                 & $[2 , 8 , 16 , 16]$                     \\
\multicolumn{1}{c|}{}                             & $E_{latent}$                                                              & $[64 , 8 , 50 , 32]$                \\
\multicolumn{1}{c|}{}                             & $D_{latent}$                                                              & $[64 , 1024 , 512 , 1024 , 2]$     \\ \hline
\end{tabular}
\caption{\label{tab:subnectworks}The architecture of the sub-networks employed in our experiments.}
\end{table}

\subsection{Time Performance}
We test our model's inference time on the SDD dataset. When sampling 20 points at each timestep for predicting one future trajectory, our model takes 336.62 seconds (0.73 seconds/scene) for the whole test dataset, 6.73s for coarse prediction, 48.50s for calculating the interaction energy maps, and 274.08s for performing the CVAE. The major overhead comes from the energy-based model and the CVAE, as the two are performed for each pedestrian at each timestep. Due to the high dimension of the energy map ($100\times100$), the CVAE constitutes the main computation complexity, which in expectation takes longer inference time than other baselines \cite{yue2022nsp, zhou2023csr}, as their CVAE is only conditioned on the past trajectories.




